\documentclass[11pt,a4paper,onecolumn]{personalarxiv}
\usepackage{fontawesome5}

\usepackage{amsmath,amsfonts,bm}

\def\eqref#1{equation~\ref{#1}}

\def\1{\bm{1}}

\DeclareMathAlphabet{\mathsfit}{\encodingdefault}{\sfdefault}{m}{sl}
\SetMathAlphabet{\mathsfit}{bold}{\encodingdefault}{\sfdefault}{bx}{n}

\usepackage{algorithm}
\usepackage{algpseudocode}
\usepackage{array,longtable, float}
\usepackage{tikz}
\usetikzlibrary{arrows.meta, positioning, fit, backgrounds, calc}

\newcolumntype{Y}{>{\raggedright\arraybackslash}X}
\newcolumntype{P}[1]{>{\raggedright\arraybackslash}p{#1}}
\newcommand{\ours}{RoboHarness-Zero}

\renewcommand{\headerfont}{
  \normalfont\color{black}\fontsize{8}{10}\selectfont
}

\title{Body and Soul: Decoupling Robotic Brain and Policy with RoboHarness-Zero}
\papershorttitle{RoboHarness-Zero: Upgrade the Policy, Keep the Harness}

\author[1,2]{Yihao Shen}
\author[1,3]{Yuxuan Chen}
\author[1,3]{Xuantong Liu}
\author[4]{Jun Gao}
\author[5]{Weijian Luo\thanks{Corresponding author: \href{mailto:pkulwj1994@icloud.com}{\texttt{pkulwj1994@icloud.com}}}}
\author[1,2]{Tianyang Hu\thanks{Corresponding author: \href{mailto:hutianyang@cuhk.edu.cn}{\texttt{hutianyang@cuhk.edu.cn}}}}

\affil[1]{Polyadic Research Group}
\affil[2]{SAGA Lab, The Chinese University of Hong Kong, Shenzhen}
\affil[3]{The Hong Kong University of Science and Technology}
\affil[4]{NVIDIA}
\affil[5]{Dots Studio, Xiaohongshu}

\paperpdfauthor{
  Yihao Shen, Yuxuan Chen, Xuantong Liu, Weijian Luo, Tianyang Hu
}

\DeclareUnicodeCharacter{03C0}{\ensuremath{\pi}}

\renewenvironment{abstract}{%
  \par\begingroup
  \normalfont\normalsize
  \setlength{\parindent}{0pt}
  \begin{center}{\large\scshape Abstract}\end{center}
  \begin{adjustwidth}{0.09\textwidth}{0.09\textwidth}
  \noindent\ignorespaces
}{%
  \par\end{adjustwidth}
  \endgroup\bigskip
}

\newcolumntype{C}[1]{>{\centering\arraybackslash}p{#1}}

\begin{document}

\maketitle

\begin{abstract}
Vision-language-action (VLA) policies can execute individual skills but fail to complete tasks that require combining several of them. In our audit of Xiaomi-Robotics-1, an estimated 43\% of failures involve choosing the wrong target or losing track of task progress. Wrapping a frozen VLA in a vision-language model (VLM) harness is a natural remedy. Existing harnesses can gain substantially from task- and policy-specific preparation, raising the question of how much of that gain transfers to a new task or policy. An ideal harness should (i) need minimal preparation for a new task, (ii) need no redesign for a new policy, and (iii) preserve standalone capabilities the policy already has. To the best of our knowledge, our \ours{} is the first harness built this way. Its premise is that choosing targets and tracking progress are easy for a VLM such as GPT-6 without task preparation, while the frozen VLA already handles continuous, contact-rich motion. The frozen VLA therefore executes continuous control while the VLM plans zero-shot and monitors execution; when progress stalls or targets drift, the VLM first rephrases the instruction and, only as a fallback, executes short generated code over generic motion primitives before returning fine control to the VLA.
The supervision rules stay fixed across policies, requiring only a thin interface adapter and an automatically profiled skill configuration per policy. On 200 RoboCasa365 scenes, \ours{} with GPT-6 Astra raises the success rate of frozen Xiaomi-Robotics-1 from \textbf{\color{blue}57.5\%} to \textbf{\color{blue}70.5\%} and preserves \textbf{\color{blue}90.4\%} of its standalone successes (104/115) while rescuing 37 of 85 failures. On a shared 50-scene subset, the same rules add 8 to 10 percentage points to each of four other frozen policies, while existing preparation-based harness systems often fail to transfer to a new policy.

\begin{center}
\small
\textcolor[HTML]{6E40C9}{\faGithub}\enspace
\textbf{Code:}\enspace
\href{https://github.com/polyadic-research/roboharness0}%
     {\nolinkurl{github.com/polyadic-research/roboharness0}}\\[0.4em]
\textcolor[HTML]{1A73E8}{\faGlobe}\enspace
\textbf{Project page:}\enspace
\href{https://odessey159.github.io/roboharness-zero/}%
     {\nolinkurl{odessey159.github.io/roboharness-zero/}}\\[0.4em]
\textcolor[HTML]{0F9D8A}{\faUsers}\enspace
\textbf{Team page:}\enspace
\href{https://home.polyadiclabs.com/}%
     {\nolinkurl{home.polyadiclabs.com/}}
\end{center}

\end{abstract}

\section{Introduction}

\label{sec:introduction}

\begin{figure}[t]
  \centering
  \includegraphics[width=\linewidth]{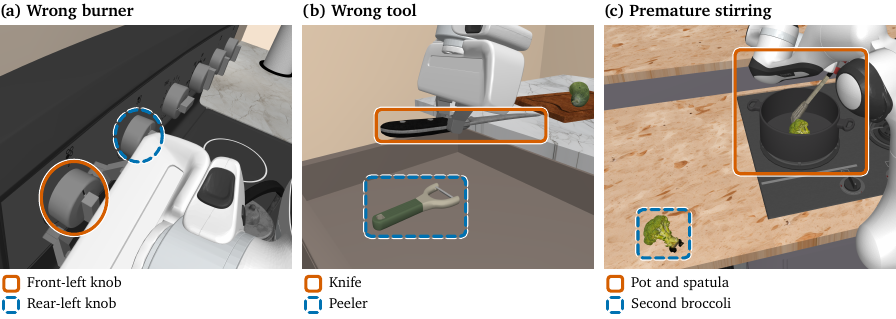}
  \caption{Target selection and sequencing errors by Xiaomi-Robotics-1 in RoboCasa365.
    \textbf{(a)} Asked to turn on the rear-left burner, the robot turns the front-left knob.
    \textbf{(b)} Asked to place a tool for peeling a cucumber on the cutting board, the robot lifts a knife while the peeler remains in the drawer. The harness later guides the gripper to the peeler, and the policy completes the task.
    \textbf{(c)} Asked to put two pieces of broccoli in a pot and then stir, the robot begins stirring while the second piece is still on the counter.
    The standalone runs in (a) and (c) both ended in failure; with the harness, both scenes succeeded.
    Solid outlines mark the objects involved in the observed action. Dashed outlines mark the intended target at that step.
    Panels (a) and (b) show external replay views unavailable to the policy. Panel (c) shows a cropped view from the policy's left camera.
    }
  \label{fig:failure-cases}
\end{figure}

A vision-language-action (VLA) policy can have the motor skills a task requires and still fail to finish it~\citep{hivla2026}. In some RoboCasa365~\citep{robocasa365} episodes, a robot acted on the wrong object, turning the wrong burner knob or taking the wrong item from a drawer (Figure~\ref{fig:failure-cases}a,b). Even when it chose the right item, it could move on too soon. Asked to put two pieces of broccoli in a pot and stir them, it added one and began stirring while the other was still on the counter. Across failed RoboCasa365 episodes with Xiaomi-Robotics-1~\citep{xiaomirobotics2026}, we estimate that 43\% of failures arise from choosing the wrong target or losing track of which steps remained (\S\ref{sec:failures}, Table~\ref{tab:failure-categories}).

A natural response to these decision errors is to put a vision-language model (VLM) in charge of tracking the task while the VLA executes it. Existing harnesses improve performance by learning how to guide the robot in a particular setting. 
In its few-shot RoboCasa365 evaluation, Harness VLA~\citep{harnessvla2026} finds a successful sequence of subtask instructions on one reference scene per task and reuses it in new scenes. EmbodiedRSI~\citep{embodiedrsi2026} runs robot trials to decide which changes to its code and skills to keep, while AdaHVLA~\citep{adahvla2026} revises its prompting and recovery code separately for each robot and policy. \textit{How much of the resulting gain survives when the task set or the policy changes?} Although Harness VLA gains 29.2 percentage points on RLDX-1, transferring its released task memories to Xiaomi-Robotics-1 on our 50-scene benchmark solves 29 scenes, compared with 30 for Xiaomi alone (\(-2.0\) percentage points).

\begin{figure}[t]
  \centering
  \includegraphics{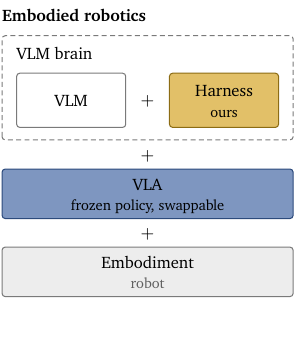}\hspace{4mm}%
  \includegraphics{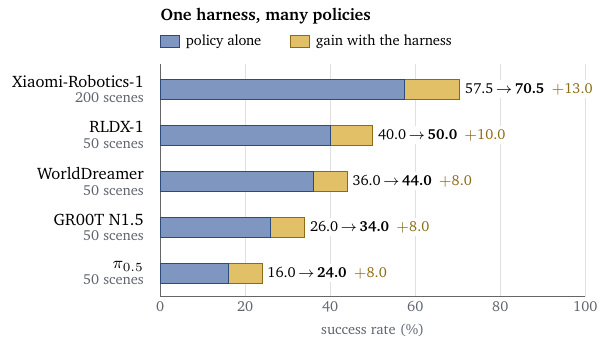}
  \caption{\textbf{Left:} RoboHarness-Zero uses a VLM to supervise a frozen VLA that controls the robot.
  \textbf{Right:} Success rates on RoboCasa365 with and without the harness.
  Xiaomi-Robotics-1 is evaluated on 200 scenes. The other four policies share
  a 50-scene subset with one scene per task. Gold bars show the net gain in percentage points.}
  \label{fig:policy-transfer}
\end{figure}

We argue that an ideal harness should \textit{(i) need minimal preparation for a new task, (ii) need no redesign for a new policy, and (iii) avoid breaking what the policy already solves}. We introduce \textbf{\ours{}}, a VLM harness that gives a frozen VLA the full task and monitors progress within each substage during execution, without interfering while the policy advances. 
When progress stalls or the robot acts on the wrong target, \ours{} first revises the instruction and, if needed, briefly takes over the arm via VLM-generated code over generic motion primitives before handing control back.
Our core insight is that the instruction and the scene alone determine which object the robot should act on, what must happen next, and whether the current step is complete, so a VLM can judge them zero-shot and identically across policies.
This allows \ours{} to supervise different tasks and frozen VLA backbones during execution with a single fixed set of rules; onboarding a new policy requires no task-specific demonstrations or scripts—only a thin interface adapter and a one-time automatic calibration (\S\ref{sec:method}).

Our method builds on the premise that a VLM can interpret an instruction, plan its steps, and detect execution drift zero-shot, while the frozen VLA provides reactive, contact-rich motor control without needing task-specific scripts or skills.
When asked to place a suitable tool for peeling a cucumber on a cutting board (Figure~\ref{fig:failure-cases}b), \ours{} gives the frozen VLA the full instruction while the VLM forms its own step-by-step plan from the robot's camera views. Judging progress against the active step, the VLM sees that reaching the drawer completes the approach, but the grasp stalls while the peeler remains inside. It first intervenes through language—``Place the green-handled peeler on the cutting board''—and only when the VLA grasps and delivers the knife instead does the harness invoke generic motion primitives via generated code on the fly to position the gripper at the peeler, before returning control to the VLA to finish (\S\ref{sec:language}--\ref{sec:recovery}).

Task-level completion criteria can be shared across policies, whereas execution reliability and timing vary. 
For example, if a task requires two pieces of broccoli in a pot, placing only one leaves the goal incomplete regardless of the policy (Figure~\ref{fig:failure-cases}c). Automatic profiling estimates each policy’s skill success rates and completion times, allowing the same supervision rules to accommodate different execution speeds.

Across 200 RoboCasa365 evaluation scenes spanning 50 tasks, \ours{} with GPT-6 Astra raises the success rate of frozen Xiaomi-Robotics-1 from 57.5\% to 70.5\% (Figure~\ref{fig:policy-transfer}). It preserves 104 of the 115 episodes the policy solves on its own (90.4\%) and rescues 37 of its 85 failures, for a net gain of +26 episodes; multi-step tasks account for +23 of this net gain, including +13 on previously unseen task compositions. On a shared 50-scene subset, the same rules add 8 to 10 percentage points to each of four other frozen policies, from \(\pi_{0.5}\) (16\% to 24\%) to RLDX-1 (40\% to 50\%). Recovery remains a bottleneck: in a separate trajectory audit, 22 of 35 failed skill attempts still do not reach their goal after the harness intervenes (\S\ref{sec:failure-analysis}).

In summary, our core contributions are:
\begin{itemize}
    \item \ours{}, a VLM harness that supervises a frozen VLA during execution; its rules need no task-specific demonstrations or scripts and are calibrated to a new policy automatically (\S\ref{sec:method}).
    \item A paired evaluation protocol, with a no-VLM control, that separates the successes a harness preserves from the failures it rescues (\S\ref{sec:protocol}).
    \item Evidence that the same rules improve five frozen policies (Xiaomi-Robotics-1, RLDX-1, WorldDreamer, GR00T N1.5, and \(\pi_{0.5}\)) on the same scenes, while retaining 90.4\% of Xiaomi-Robotics-1's standalone successes (\S\ref{sec:main-results}--\ref{sec:many-policies}).

\end{itemize}

\section{Related Work}

\label{sec:related}

\subsection{VLM harnesses for robot manipulation}

Language models can guide manipulation through a learned hierarchy, as in Steerable VLA~\citep{steerablevla2026}, Hi Robot~\citep{hirobot2025}, and \(\pi_{0.5}\)~\citep{pi052025}, or by calling robot skills and motion primitives directly, as in SayCan~\citep{saycan2023}, Code as Policies~\citep{codeaspolicies2022}, Show-Harness~\citep{showharness2026}, Robo-Harness K1~\citep{roboharnessk12026}, and HexaShell~\citep{hexashell2026}. A third approach places a harness around a frozen action policy. We relate these harnesses to the three properties of an ideal harness (\S\ref{sec:introduction}).

Most harnesses need preparation before they can be used on a task, contrary to property~(i). On RoboCasa365, Harness VLA~\citep{harnessvla2026} explores one reference scene per task and stores a successful sequence of primitives that its planner can reuse, calling the VLA for contact-rich steps. HarnessPAI~\citep{harnesspai2026} instead uses a coding agent to revise a separate control program for each task between trials. SERA~\citep{sera2026} evaluates harness revisions through environment rollouts, while EmbodiedRSI~\citep{embodiedrsi2026} uses robot trials to select changes to code and skills. Cracking Mobile Manipulation~\citep{crackingmobile2026} uses demonstrations to guide planning and uses separate interventions to correct base positioning and local manipulation. 
Preparation also underlies the top RoboCasa365 leaderboard entry,  HexaShell~\citep{hexashell2026}, which represents a separate, policy-free line of work where a VLM controls the robot directly by writing code over motion and perception tools.
Drafting a one-page skill from earlier runs for each of the 50 tasks—recording judging criteria, step flow, and measured parameters—lifts HexaShell from 59.3\% with no task-specific input to 85.6\%; by contrast, our work focuses on harnessing existing frozen VLAs without per-task preparation.

Few harnesses carry one design across policies, as property~(ii) requires. AdaHVLA~\citep{adahvla2026} specializes its coordination rules to each robot--executor pair. On RoboCasa365, PhyAgentOS~\citep{phyagentos2026} applies a common verifier to three frozen policies after the initial rollout fails, at which point it can issue a recovery instruction from the robot's current state. Because it waits for the episode to fail, it does not put the policy's existing successes at risk, but it also cannot act while an error is still reversible.

Our harness gives the frozen policy the full task instruction and leaves it in control by default, while the VLM monitors each substage during execution. If progress stalls or the robot selects the wrong target (Figure~\ref{fig:failure-cases}), the VLM can intervene while the error is still reversible. A one-time calibration lets the same supervision rules transfer to a new policy without retraining, per-task demonstrations, or exploratory rollouts on the evaluation tasks. To our knowledge, none of the harnesses above separates, on paired runs, the successes it preserves from the failures it rescues; we report both (\S\ref{sec:protocol}).

\subsection{When and how to intervene}

When should a harness intervene, and what should it change? Building on earlier work on language feedback~\citep{innermonologue2022,doremi2024}, Mostly Harmless~\citep{mostlyharmless2026} learns to rewrite instructions and retains the original when a calibrated predictor finds little evidence that rewriting will help.

Many recovery methods act directly on motion. ForkCast~\citep{forkcast2026} selects among continuing, changing viewpoint, and replanning the arm trajectory. CheckVLA~\citep{checkvla2026} verifies actions with a trained world model, while IDR~\citep{idr2026} refines actions across several frozen policies without training. IDR adjusts every action the policy produces; our harness leaves the policy's actions untouched unless a substage stalls or the target is wrong, and gives an off-the-shelf VLM responsibility for deciding when and how to intervene. Tracking the current substage against the full instruction lets it recognize a precise grasp of the wrong object, revise the instruction first, and escalate to a bounded takeover when needed.

RESCUE~\citep{rescue2026} compares alternatives from the same checkpoint and shows how unnecessary intervention can disrupt success. Hi-VLA~\citep{hivla2026} finds that reliable controllers still benefit from orchestration. StageCraft~\citep{stagecraft2026} finds that stronger policies need fewer scene adjustments. These findings are compatible: selective intervention can preserve existing successes while timely correction can rescue failures. Our evaluation measures both effects separately, as retention and rescue (\S\ref{sec:protocol}).

\subsection{Failure detection}

SAFE~\citep{safe2025} predicts failure on unseen tasks from a VLA's internal features, while external monitors judge success from images or learn from teleoperator interventions to flag mistakes~\citep{du2023successdetectors,valueformer2026}. A detector can flag repeated unsuccessful grasps as they occur, but the alarm alone leaves open how to recover~\citep{failurespot2026}. RESCUE~\citep{rescue2026} shows that alarms can arrive too late or prompt unnecessary intervention, and Soddy et al.~\citeyearpar{soddy2026predictable} predict MemoryVLA failures early yet find that none of four interventions reliably improves success. Our VLM judges progress on the current substage from images and robot state without per-policy detector training, and our analysis of what follows an alert likewise identifies recovery as the bottleneck.

\section{\ours{}: A Fixed Harness for Any Frozen Policy}

\label{sec:method}

\subsection{VLAs fail on decisions, not only motions}

\label{sec:failures}

We audited 193 of the 893 failed rollouts from an earlier RoboCasa365 evaluation of Xiaomi-Robotics-1 running without a harness, sampled across five task groups (Appendix~\ref{app:failure-audit}).

After weighting by the number of failures in each task group, we estimate that 43\% of failures involve a decision error. The robot chooses the \emph{wrong target} in 24\% of failures and loses track of task order or progress in 19\% (an \emph{order or progress error}). The remaining 57\% comprise physical execution failures (53\% of all failures), episodes that exhaust their step budget while still making progress (3.7\%), and violations of a task constraint (0.3\%). We distinguish two types of execution failure according to whether the policy has ever completed the same task in another trial: \emph{inconsistent execution} when it has, and \emph{execution with no observed task success} when it has not. Table~\ref{tab:failure-categories} in the appendix gives the definition and an example of each category.

These findings motivate our focus: using a VLM to ground the policy’s decisions in the instruction and the current scene. The VLM does so both when selecting the intended target and when judging whether the robot is ready to move on to the next step.

\subsection{Overview and design principles}

\label{sec:setup}

\begin{figure}[t]
  \centering
  \includegraphics[width=\linewidth]{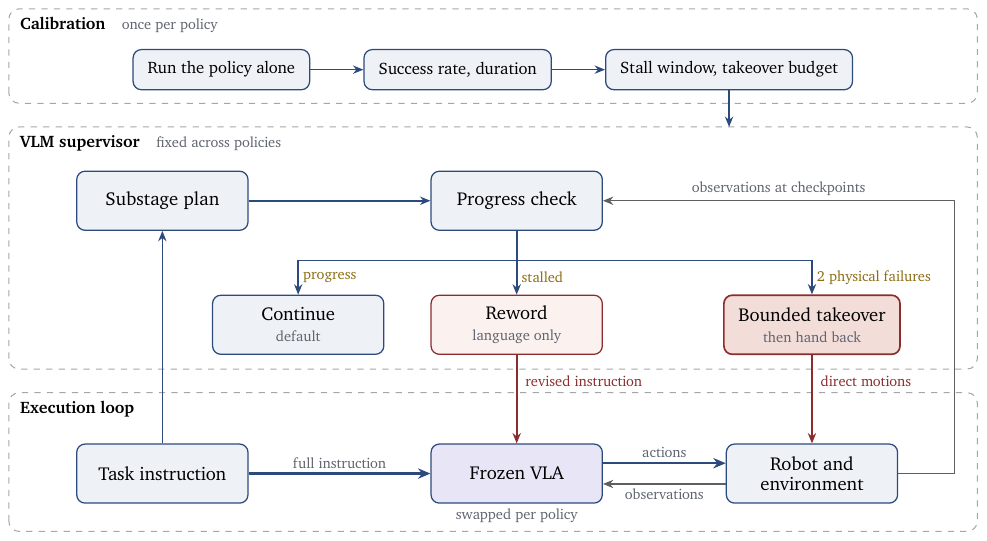}
  \ifnum\value{figure}=0\relax\stepcounter{figure}\fi
  \caption{How \ours{} divides control between a frozen VLA and a supervising VLM. \textbf{Execution loop:} the VLA receives the full task instruction and controls the robot by default. \textbf{VLM supervisor:} at each checkpoint, the VLM checks progress on the current substage. If the substage advances, the policy continues. If it stalls, the VLM rewords the instruction and the policy keeps control. After two verified physical failures, the VLM can take over for a bounded number of steps before handing control back. \textbf{Calibration:} a profiling run of the policy alone sets how long the VLM waits and how long a takeover may last. Only the VLA, its adapter, and this calibration change with the policy.}
  \label{fig:harness-architecture}
\end{figure}

Figure~\ref{fig:harness-architecture} illustrates how \ours{} divides control between a frozen VLA and a supervising VLM. The VLA receives the task instruction and generates robot actions in short sequences. At checkpoints, the VLM sees the left, right, and wrist camera views, the poses of the base and end effector, and the gripper positions. It interprets these observations in light of the original instruction, its current plan, and the commands executed since its last review. When the environment detects that the whole task has succeeded, the harness ends the episode at once and records a success. Before that, it receives no success signal. Intermediate judgments, such as whether the robot has grasped the intended object, must be made from the available observations. The robot interface supplies no ground-truth object poses, hidden fixture states, substage completion labels, or expert trajectory.

When the evidence is ambiguous, the VLM can call perception tools—or short generated code—to inspect a fresh view, zoom in, or back-project depth at a visible pixel into 3D. If the policy is making progress, it continues under the same instruction. If the VLM identifies a wrong target or a stalled step, it first revises the instruction and observes how the policy responds. When a takeover is warranted, the harness authorizes a limited sequence of base, torso, arm, or gripper motions, which consume the same episode budget as policy actions; the VLM then checks the result in a fresh observation and returns control to the VLA.

\ours{} is designed around the three properties of an ideal harness (\S\ref{sec:introduction}). For property~(i), minimal preparation for a new task, the VLM forms its own plan from the instruction and the scene in each episode, so the harness uses no task-specific demonstrations or scripts (\S\ref{sec:capability}). For property~(ii), no redesign for a new policy, the supervision rules stay fixed and a new policy brings only a native adapter and a skill profile (\S\ref{sec:calibration}). Property~(iii), breaking nothing the policy already solves, depends on when and how the VLM intervenes.

The choice of when to intervene matters because supervision can change outcomes in either direction. On paired episodes, let \(b\) denote the success rate of the standalone policy. \emph{Retention} \(r\) is the fraction of its successes retained with the harness, and the \emph{rescue rate} \(s\) is the fraction of its failures that succeed with the harness. The harness succeeds at rate \(S_H=br+(1-b)s\). For \(0<b<1\), matching or exceeding the standalone policy requires

\begin{equation}
\label{eq:retention}
S_H \ge b
\quad\Longleftrightarrow\quad
b(1-r)\leq(1-b)s
\quad\Longleftrightarrow\quad
r\geq 1-\frac{1-b}{b}s .
\end{equation}

The left side of the middle inequality counts, as a fraction of all episodes, the successes lost under supervision. The right side counts the new successes. For a given rescue rate \(s\), a stronger policy can therefore tolerate fewer losses among the episodes it already solves.

This trade-off motivates three design choices for property~(iii). First, we give the VLA the full instruction by default and use its measured skill success rates and execution times to set the initial level of supervision, giving familiar behavior time to succeed (\S\ref{sec:capability}). Second, we judge progress within the current substage, so a successful approach cannot mask a stalled grasp (\S\ref{sec:progress}). Third, we ordinarily revise the instruction before authorizing a bounded takeover, and we hand execution back after the correction (\S\ref{sec:language}--\ref{sec:recovery}). This lets the VLM redirect a failing step while leaving the rest of the task to the policy.

\subsection{Hand over the whole task, then watch}

\label{sec:capability}

\ours{} needs no preparation for a new task: every episode starts from the task instruction alone, with no demonstrations, memories of earlier episodes, or task scripts (Appendix~\ref{app:task-information}). 
At the start of a composite task with no whole-task entry in the skill profile, \ours{} passes the original instruction verbatim to the frozen VLA; most composite tasks in our evaluation fall under this rule, while tasks matching a profiled entry adjust only the initial check interval and takeover permissions according to measured reliability (Appendix~\ref{app:task-information}).
The VLM uses its own plan to identify what would count as progress at each substage. Reaching a new substage updates what the VLM watches for, while the policy continues under the original instruction.

Monitoring takes place between windows of policy execution. Xiaomi-Robotics-1 generates 16 actions per prediction. A default window allows ten predictions, or 160 control steps (eight seconds of simulated motion). At a checkpoint, the VLM can inspect fresh left, right, and wrist views alongside the robot state and the actions executed since its last review. Its default response is to let the policy continue, without sending a new instruction or motor command (Appendix~\ref{app:development-cases} gives an example).

If the VLM issued the steps one at a time, each transition between steps would depend on its judgment. Keeping the substage plan on the monitoring side lets the policy carry out that transition itself while the VLM checks the result (Appendix~\ref{app:development-cases}).

\subsection{Progress is judged at the current substage}

\label{sec:progress}

While the policy executes, the VLM asks one question at each checkpoint: has the robot made progress on the current substage? The VLM's plan divides the task into substages and each substage into one or more phases, such as approaching an object and then grasping it. Each phase has a target, the object or location it acts on. In the peeler episode of Figure~\ref{fig:failure-cases}b, reaching the utensils completes the approach phase, but the grasp phase stays incomplete while the peeler remains below the counter edge. 
The harness records progress for the current phase and its target. The first 3D distance estimate from the end effector to the depth-localized target establishes a reference. A fresh observation earns progress credit when the estimated distance falls at least 2 cm below its previous minimum, or when the VLM first verifies that the phase is complete. Once the robot has grasped an object, seeing it still in the gripper confirms that the grasp is secure. Progress toward placement requires fresh evidence about the object’s destination. Otherwise, credit for holding the object could hide a stalled placement (Appendix~\ref{app:development-cases}).

The VLM can continue, advance, or review. While the current phase is underway, the policy continues under its existing instruction. Verified progress resets the stall counter. Otherwise, the harness adds to the counter the number of control steps the policy has executed since the previous checkpoint. When the VLM verifies phase completion, it advances to the next phase or, when the substage has none left, to the next unfinished substage in its plan. At any checkpoint where the counter reaches the policy's stall window (\S\ref{sec:calibration}), the harness raises a \emph{stall alert}: the VLM reviews the remaining goal and may give the policy a more specific instruction, while leaving low-level motor execution with the policy rather than taking over. For Xiaomi-Robotics-1 the stall window is 120 steps, less than a full 160-step window, so one full window without verified progress triggers a review. The policy retains motor control during this review. RoboCasa’s task-level success predicate determines when the full task has succeeded.

\subsection{Reword before taking over}

\label{sec:language}

When a task stalls, the first intervention available to the VLM is to change the instruction. The VLM can narrow a multi-step request to the step where the robot is stuck, rephrase the same goal as a short imperative, or identify the intended object by its visible color, shape, or location. In the peeler episode, the VLM tells the policy, ``Place the green-handled peeler on the cutting board,'' which names the tool to pick up. The frozen VLA still chooses how to approach, grasp, and move it. This lets the VLM correct the policy’s interpretation of the scene through the policy’s existing language interface, without access to its internal state or action representation.

Escalation depends on what happens when the robot acts. Two verified physical failures on the same substage, such as two missed grasps, make a bounded takeover available. The harness carries this count across instruction changes. Repeated rewording alone never triggers a takeover. A dropped object, an unsafe motion, or a confirmed conflict with the task, such as grasping the wrong object, can justify an earlier takeover. Section~\ref{sec:gain} reports which paired gains involve instruction rewriting alone. Section~\ref{sec:ablations} compares the full harness with a no-VLM control that periodically reissues the original instruction.

\subsection{Bounded takeover and hand-back}

\label{sec:recovery}

Once takeover is available (\S\ref{sec:language}), the VLM can reposition the base, adjust torso height, and guide the gripper through waypoints identified from camera and depth observations. In the peeler episode, the policy delivers the knife even after the reworded instruction. The harness then guides the gripper to the peeler and returns control to the policy, which completes the task. For skills the policy has demonstrated it can perform, the VLM brings the hand into position and leaves fine contact to the policy. For weak or unmeasured skills, the harness also permits direct grasping, placement, and pouring. Closing the gripper on an object requires a fresh wrist-camera check, and the VLM must verify after lifting that the object moved with the hand.

A takeover draws from the same episode budget as the policy. Its initial allowance scales with the skill’s measured typical duration, with a fixed default when no measurement is available. The harness ends the attempt when it stops making progress or upon a verified drop or unsafe motion. If the robot is already moving a securely held object toward placement, it can receive a short, capped extension (Appendix~\ref{app:constants}). To hand control back, the VLM positions the base and hand so the target is visible and within reach. It preserves any held object, reissues the instruction for the current substage, and checks progress after a short window of fresh policy actions. Section~\ref{sec:gain} reports how often takeover is used, and Section~\ref{sec:failure-analysis} examines where it fails.

\subsection{Calibrating to a new policy}

\label{sec:calibration}

Switching policies leaves the rules unchanged, as property~(ii) requires. It needs only a native adapter and a set of measured constants (Table~\ref{tab:policy-calibration}). 
The adapter maps camera images, robot state, and the active instruction into the policy's input format and returns action chunks through a common interface. 
Profiling supplies three policy-specific quantities used by the fixed rules: per-skill success rates (which govern whether takeover may grasp or place directly; \S\ref{sec:recovery}), typical skill durations (which scale takeover budgets; \S\ref{sec:recovery}), and stall windows (how many steps the policy may execute without verified progress before triggering a review; \S\ref{sec:progress}). Xiaomi-Robotics-1 served as our development policy, for which we fixed a global 120-step stall window during development; all four transfer policies use stall windows derived automatically from profiling with zero manual tuning.

We profile a frozen policy by running it alone on 65 atomic RoboCasa tasks on separate profiling seeds, with 15 episodes per task. Each episode runs until the environment reports success or the task horizon expires. For skill \(k\), we estimate its success rate as the fraction of its 15 trials that succeed and its typical duration as the median step of first success among those trials. The 975 rollouts take 6.16 cumulative worker-hours for GR00T N1.5, with zero VLM calls.

To set a policy’s stall windows, we use its successful profiling trajectories to estimate how long a policy that eventually succeeds can appear to make no progress. For each trajectory \(i\), let \(g_i\) be the longest interval, in control steps, during which an offline measure of end-effector motion registers no progress. For each skill \(k\), we set \(W_k=\max(1,\lceil 1.5\,Q_{0.95}(\{g_i:i\in k\})\rceil)\), using a fixed multiplier of 1.5. When a substage has no matching skill, we use the same calculation across all successful trajectories. During execution, the VLM continues to judge progress from evidence about the current target and substage, as described in \S\ref{sec:progress}. Each policy uses its own measured skill success rates and durations. Appendix~\ref{app:stall-windows} lists the stall windows used for each policy.

The fallback matters on unfamiliar task compositions. In the Sol trajectory audit (\S\ref{sec:protocol}), only 16 of 261 distinct registered substages (6.1\%) match a measured skill. The other 245 have no matching entry. A policy-wide stall window therefore lets the harness adapt how long it waits even when it cannot identify a previously measured skill.

\begingroup
\fontsize{9.2}{11.2}\selectfont
\setlength{\tabcolsep}{4pt}
\renewcommand{\arraystretch}{1.14}
\begin{longtable}{@{}P{0.22\linewidth}P{0.72\linewidth}@{}}
\caption{What stays fixed, what is replaced, and what is measured when \ours{} is moved to a new frozen policy.}\label{tab:policy-calibration}\\
\toprule
\textbf{Status across policies} & \textbf{Components} \\
\midrule
\endfirsthead
\toprule
\textbf{Status across policies} & \textbf{Components} \\
\midrule
\endhead
\bottomrule
\endfoot
\bottomrule
\endlastfoot
Fixed & Observation and action interface exposed to the harness, substage progress criterion, order of language and motor intervention, takeover guards, and hand-back rules \\
\addlinespace[4pt]
Replaced & Frozen policy and the adapter that translates between its native inputs and outputs and the common interface \\
\addlinespace[4pt]
Measured or derived & Per-skill success rates and typical durations. Per-skill stall windows and a pooled fallback, derived from the policy's profiling or left at default values \\
\addlinespace[4pt]
\end{longtable}
\endgroup

\section{Experiments}

\label{sec:experiments}

\subsection{Setup and protocol}

\label{sec:protocol}

We evaluate the 50 tasks of the RoboCasa365 leaderboard (18 atomic-seen, 16 composite-seen, and 16 composite-unseen)~\citep{robocasa365,robocasaleaderboard2026}. Seen and unseen refer to whether a task appears in the policies' training data. Four scenes per task (episode indices 1–4, environment seed 7) yield 200 episodes scored by the environment’s success predicate. Table~\ref{tab:episode-sets} lists every episode set used in this paper. Our 72 development episodes cover 47 of these tasks. Ten evaluation scenes match development episodes on the initial simulator state and all three camera images. Excluding them gives 134/190 harness successes versus 108/190 standalone successes.

\begin{table}[htbp]
\centering
\caption{Episode sets used in this paper. Only the main evaluation and its shared subset report performance. The others support failure analysis, development, and calibration. Appendix~\ref{app:episode-sets} gives their overlap.}
\label{tab:episode-sets}
\small
\setlength{\tabcolsep}{4pt}
\begin{tabular}{@{}lp{0.44\linewidth}p{0.28\linewidth}@{}}
\toprule
Name & Episodes & Used for \\
\midrule
Failure audit & 193 of the 893 failures in an earlier evaluation of Xiaomi-Robotics-1 alone & Failure categories (\S\ref{sec:failures}) \\
Development episodes & 72 Xiaomi episodes covering 47 tasks & Developing the harness \\
Main evaluation & 200 scenes: 50 tasks, episode indices 1--4 & Main results (\S\ref{sec:main-results}, \S\ref{sec:gain}--\ref{sec:failure-analysis}) \\
Shared 50-scene subset & Episode index 1 of each main-evaluation task & Other policies and Harness VLA (\S\ref{sec:many-policies}--\ref{sec:harness-vla-transfer}) \\
Sol trajectory audit & 80 unseen-composition episodes supervised by GPT-6.1 Sol: 16 tasks, episode indices 0--4 & Skill matching and recovery (\S\ref{sec:recovery}, \S\ref{sec:calibration}, \S\ref{sec:failure-analysis}) \\
Profiling rollouts & 975 per policy, run alone: 65 atomic tasks, 15 episodes each & Skill success rates and durations (\S\ref{sec:calibration}) \\
\bottomrule
\end{tabular}
\end{table}

\noindent\textbf{Paired protocol.} 
Paired rollouts share the initial simulator state, initial policy random-number state, and control horizon. Trajectories can diverge after intervention. Retention measures the fraction of standalone successes preserved with the harness. Rescue rate measures the fraction of standalone failures that succeed with it. Our no-VLM control, \emph{fixed reissue}, reissues the original instruction every 160 control steps. We use a two-sided exact McNemar test and a permutation test that swaps method labels jointly by task.

We pair frozen Xiaomi-Robotics-1~\citep{xiaomicheckpoint2026} with the harness (Appendix~\ref{app:evaluation-protocol} gives versions and timeouts). Xiaomi executes 16-action chunks at 20 Hz. The default monitoring interval is 160 steps and the stall window is 120 steps. The four other policies use their native adapters (\S\ref{sec:many-policies}). For the main Xiaomi evaluation, we queried gpt-6-astra at xhigh reasoning effort. We also evaluated GPT-6.1 Sol (\texttt{gpt-6.1-sol}) at medium reasoning effort as an alternative VLM.

During each episode, the VLM has access to the task instruction, images from the left, right, and wrist cameras, the robot state, and feedback from executed commands. It can query depth at a selected image point to locate the corresponding point in 3D (\S\ref{sec:setup}). We provide no demonstrations, scripted solutions, or separate exploration episodes.

We evaluate Harness VLA with its released per-task memories on the same 50 scenes used for policy transfer (\S\ref{sec:harness-vla-transfer}). The 200 harness runs in the main result average 71 VLM requests and 1,136 control steps per episode. Standalone runs average 1,327 steps.

\subsection{Main results on RoboCasa365}

\label{sec:main-results}

How many additional scenes does the harness help a strong frozen policy complete, and how many of its original successes does it lose? Across 200 paired RoboCasa365 scenes, Xiaomi-Robotics-1 succeeds on 115 alone (57.5\%) and on 141 with Astra (70.5\%). The standalone rate is close to the 57.4\% that the RoboCasa365 leaderboard reports for this policy~\citep{robocasaleaderboard2026}, so our scenes give a baseline in line with the official evaluation. The gain is 13.0 percentage points, with a 95\% paired bootstrap confidence interval of 6.5–19.5 points. With Sol supervising the same policy, the harness solves 126 scenes (63.0\%), a gain of 5.5 points. Reissuing the original instruction every 160 steps without a VLM yields 113/200 successes (56.5\%), one percentage point below Xiaomi alone (Table~\ref{tab:success-by-category}; \S\ref{sec:ablations}).

\begingroup
\fontsize{9.2}{11.2}\selectfont
\setlength{\tabcolsep}{4pt}
\renewcommand{\arraystretch}{1.14}
\begin{longtable}{@{}P{0.225\linewidth}C{0.175\linewidth}C{0.175\linewidth}C{0.175\linewidth}C{0.175\linewidth}@{}}
\caption{Success by task category. Entries give successes/episodes and success rate in percent.}\label{tab:success-by-category}\\
\toprule
\textbf{Method} & \textbf{Atomic-seen} & \textbf{Composite-seen} & \textbf{Composite-unseen} & \textbf{Overall} \\
\midrule
\endfirsthead
\toprule
\textbf{Method} & \textbf{Atomic-seen} & \textbf{Composite-seen} & \textbf{Composite-unseen} & \textbf{Overall} \\
\midrule
\endhead
\bottomrule
\endfoot
\bottomrule
\endlastfoot
Xiaomi alone & 59/72, 81.9 & 35/64, 54.7 & 21/64, 32.8 & 115/200, 57.5 \\
\addlinespace[4pt]
Xiaomi + fixed reissue (no VLM) & 56/72, 77.8 & 33/64, 51.6 & 24/64, 37.5 & 113/200, 56.5 \\
\addlinespace[4pt]
Xiaomi + harness (Sol) & 59/72, 81.9 & 43/64, 67.2 & 24/64, 37.5 & 126/200, 63.0 \\
\addlinespace[4pt]
Xiaomi + harness (Astra) & 62/72, 86.1 & 45/64, 70.3 & 34/64, 53.1 & 141/200, 70.5 \\
\addlinespace[4pt]
\end{longtable}
\endgroup

With Astra, the harness preserves 104 of Xiaomi’s 115 successes (90.4\% retention) and solves 37 of the 85 scenes Xiaomi misses (43.5\% rescue). The 37 gains and 11 losses yield a two-sided exact McNemar p-value of 0.00022. Composite tasks account for 23 of the 26 net additional successes. On seen compositions, the harness gains 17 scenes and loses seven. On unseen compositions, it gains 15 and loses only two, accounting for half of the overall gain. The improvement appears in all four batches of 50 scenes. In episode-index order, success counts rise from 30 to 34, 29 to 36, 31 to 34, and 25 to 37.

\begin{figure}[htbp]
  \centering
  \includegraphics[width=0.69\linewidth]{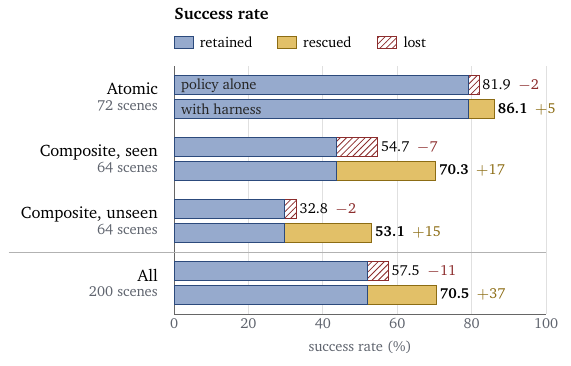}
  \caption{Success rates for Xiaomi-Robotics-1 alone and with GPT-6 Astra supervision on 200 paired RoboCasa365 scenes. Signed counts give scenes solved only with the harness ($+$) or only by the policy alone ($-$).}
  \label{fig:paired-results}
\end{figure}

Figure~\ref{fig:paired-results} shows the paired gains and losses by task category. Tables~\ref{tab:success-by-category} and~\ref{tab:paired-outcomes} report success rates and paired outcomes, respectively.

Among standalone VLAs, the RoboCasa365 leaderboard reports 62.6\% for minionx~\citep{robocasaleaderboard2026}. Among LLM–VLA systems, Harness VLA reports 59.2\% with GPT-6 Astra on RLDX-1, using one reference scene per task~\citep[arXiv v5, Appendix~H, Table~22]{harnessvla2026}, and EmbodiedRSI reports 77.0\% after exploration~\citep{embodiedrsi2026}. OpenRUA, which uses a language model with inverse kinematics, reports 39.4\%~\citep{openrua2026}. HexaShell 1.5 reports 85.6\% using Cartesian motion control and per-task skill notes~\citep{hexashell2026}. Differences in scenes, task-specific preparation, and access to success feedback preclude a direct ranking.

The harness retains nine in ten Xiaomi successes while delivering its largest gain on unseen compositions. Section~\ref{sec:gain} investigates the sources of the gain.

\subsection{One harness, many policies}

\label{sec:many-policies}

Can a harness developed with Xiaomi-Robotics-1 improve a weaker frozen policy while preserving the tasks it already solves? We tested four policies: RLDX-1~\citep{rldx2026}, WorldDreamer~\citep{worlddreamercheckpoint2026}, GR00T N1.5~\citep{grootn152025}, and \(\pi_{0.5}\)~\citep{pi052025}. GR00T N1.5 and \(\pi_{0.5}\) use the RoboCasa365 multitask checkpoints at training steps 120,000 and 75,000, respectively~\citep{robocasacheckpoints2026}, and RLDX-1 and WorldDreamer use their released RoboCasa365 checkpoints (Appendix~\ref{app:evaluation-protocol}). All four run on the shared 50-scene subset, episode index 1 of each task (Table~\ref{tab:episode-sets}). They solve 40\%, 36\%, 26\%, and 16\% of these scenes on their own, compared with 60\% for Xiaomi. Each uses its native adapter, its own skill profile, and the same rules for monitoring progress, revising instructions, and taking over (\S\ref{sec:calibration}, Table~\ref{tab:policy-calibration}). GR00T N1.5 and \(\pi_{0.5}\) share the default stall windows, while RLDX-1 and WorldDreamer use windows derived from their own profiling (Appendix~\ref{app:stall-windows}).

With Astra supervising, RLDX-1 improves from 20 to 25 successes, WorldDreamer from 18 to 22, GR00T from 13 to 17, and \(\pi_{0.5}\) from 8 to 12 (Figure~\ref{fig:policy-transfer}). On the same 50 scenes, Xiaomi improves from 30 to 34, so all five policies gain eight to ten percentage points (Table~\ref{tab:policy-transfer-pairs}). The four new policies lose one of their 59 standalone successes: GR00T loses one of 13, while RLDX-1, WorldDreamer, and \(\pi_{0.5}\) retain all of theirs. In 54 of the 58 retained successes, the harness keeps the original instruction and the policy generates every action. All 17 such episodes for GR00T and \(\pi_{0.5}\) finish on exactly the same control step as the standalone run, as do 24 of the 37 for RLDX-1 and WorldDreamer.

\begin{table}[htbp]
\centering
\caption{Paired outcomes with Astra supervision on the same 50 RoboCasa365 scenes, one per task. Retention and rescue are computed among standalone successes and failures, respectively.}
\label{tab:policy-transfer-pairs}
\small
\setlength{\tabcolsep}{4pt}
\begin{tabular}{@{}lrrrrrr@{}}
\toprule
Frozen policy & \shortstack{Both\\succeed} & \shortstack{Policy\\only} &
\shortstack{Harness\\only} & \shortstack{Both\\fail} & Retention & Rescue \\
\midrule
Xiaomi-Robotics-1 & 28 & 2 & 6 & 14 & 93.3\%  & 30.0\% \\
RLDX-1            & 20 & 0 & 5 & 25 & 100.0\% & 16.7\% \\
WorldDreamer      & 18 & 0 & 4 & 28 & 100.0\% & 12.5\% \\
GR00T N1.5        & 12 & 1 & 5 & 32 & 92.3\%  & 13.5\% \\
\(\pi_{0.5}\)     &  8 & 0 & 4 & 38 & 100.0\% &  9.5\% \\
\bottomrule
\end{tabular}
\end{table}

\enlargethispage{\baselineskip}
The gains fall in different places. All five RLDX-1 rescues are unseen compositions, a category in which RLDX-1 alone solves no scene. WorldDreamer gains one atomic task and three seen compositions, and GR00T's net gain has the same split. All four gains for \(\pi_{0.5}\) come from atomic tasks. The harness recovers 5 of 30 standalone failures for RLDX-1, 4 of 32 for WorldDreamer, 5 of 37 for GR00T, and 4 of 42 for \(\pi_{0.5}\), compared with 6 of 20 for Xiaomi on the same scenes. Takeover is involved in all five RLDX-1 rescues, all four \(\pi_{0.5}\) rescues, three of GR00T's five, and one of WorldDreamer's four. Four of the remaining rescues use instruction rewriting alone, including one in which the VLM asked GR00T to close the dishwasher door, and one WorldDreamer rescue involves no intervention. Supervision averages 162 VLM requests per episode for \(\pi_{0.5}\), about 115 for RLDX-1 and WorldDreamer, and 98 for GR00T. The rules developed with Xiaomi thus carry over: each new policy gains eight to ten points, and together they lose a single standalone success.

\subsection{Does a task-prepared harness transfer to a stronger policy?}
\label{sec:harness-vla-transfer}

Does the gain from preparing a harness for each task carry over to a stronger policy? Harness VLA is the closest public comparison on RoboCasa365. It uses a reference scene for each task to build a memory of how to solve that task. With RLDX-1, the reported gains over the 30.0\% standalone baseline are 27.2 percentage points with GPT-5.5 and 29.2 points with GPT-6 Astra~\citep[arXiv v5, Table~4 and Appendix~H, Table~22]{harnessvla2026}.

We ran the official Harness VLA code with Xiaomi-Robotics-1 and the released memories, without collecting new traces or revising the memories for Xiaomi. Both harnesses used GPT-6 Astra at \texttt{xhigh} reasoning effort on the same 50 scenes as in \S\ref{sec:many-policies}. Harness VLA solves 29 scenes, Xiaomi alone solves 30, and our harness solves 34 (Table~\ref{tab:harness-vla-transfer}). The two harnesses tie on atomic tasks. Our five additional successes come from composite tasks, three on seen compositions and two on unseen compositions. Four tasks lack task-specific notes in the released memory. Harness VLA still solves three of these scenes, and without the four it solves 26 of the remaining 46, against 28 for Xiaomi alone and 31 for our harness (Appendix~\ref{app:harness-vla-transfer}).

\begin{table}[!htb]
\centering
\caption{Transfer of Harness VLA on RoboCasa365. Panel A reproduces the RLDX-1 baseline from Table~4 and the two Codex planner results from Appendix~H, Table~22 of \citet{harnessvla2026} (arXiv v5). Panel B uses the shared 50-scene subset (Table~\ref{tab:episode-sets}); its last row repeats our main result on all 200 scenes, where Xiaomi alone scores 57.5\%. Gains are percentage-point differences from the standalone policy on the same scenes.}
\label{tab:harness-vla-transfer}
\small
\setlength{\tabcolsep}{4pt}
\begin{tabular}{@{}lrrrrr@{}}
\toprule
Policy and method & \shortstack{Atomic\\seen} &
\shortstack{Composite\\seen} & \shortstack{Composite\\unseen} &
Overall & \shortstack{Gain\\(pp)} \\
\midrule
\multicolumn{6}{@{}l}{\textbf{A. Published RLDX-1 results}} \\
RLDX-1 alone              & 60.0\% & 21.3\% & 5.0\%  & 30.0\% & --- \\
\quad + Harness VLA (GPT-5.5) & 92.2\% & 61.3\% & 13.8\% & 57.2\% & \(+27.2\) \\
\quad + Harness VLA (Astra)   & 87.8\% & 43.8\% & 42.5\% & 59.2\% & \(+29.2\) \\
\midrule
\multicolumn{6}{@{}l}{\textbf{B. Xiaomi on the shared 50 scenes unless noted}} \\
Xiaomi alone                  & 88.9\% & 50.0\% & 37.5\% & 60.0\% & --- \\
\quad + Harness VLA (Astra)   & 83.3\% & 43.8\% & 43.8\% & 58.0\% & \(-2.0\) \\
\quad + Our harness (Astra)   & 83.3\% & 62.5\% & 56.3\% & 68.0\% & \(+8.0\) \\
\quad + Our harness (Astra, 200 scenes) & 86.1\% & 70.3\% & 53.1\% & 70.5\% & \(+13.0\) \\
\bottomrule
\end{tabular}
\end{table}

On RLDX-1, the policy it was prepared for, Harness VLA's published gain of 29.2 points is about three times ours of ten (\S\ref{sec:many-policies}). The standalone baselines differ, 30.0\% in the published evaluation and 40\% on our 50 scenes, so we compare gains rather than absolute rates. On Xiaomi the order reverses. With the released memories, Harness VLA retains 26 of Xiaomi's 30 standalone successes and rescues three of its 20 failures, ending one scene below the policy alone, while our harness retains 28 and rescues six. Harness VLA loses successes and rescues failures at similar rates, 13\% and 15\%, but Xiaomi has 30 successes to lose and only 20 failures to rescue, so the losses outweigh the rescues. We did not rebuild the memories for Xiaomi: that is the per-policy preparation that property~(ii) asks a harness to do without, and this comparison measures what remains without it. The gain from per-task preparation therefore does not carry over to the stronger policy here, whereas our smaller gain holds on all five policies.

\subsection{Where the gain comes from}

\label{sec:gain}

Of the 37 gains and 11 losses reported in \S\ref{sec:main-results}, how many involve a change of instruction or a takeover, and which gains show evidence that the intervention helps?

Table~\ref{tab:intervention-by-outcome} groups the 200 paired episodes by the highest level of intervention used in the harness run. Xiaomi generates every robot action in 22 of the 37 harness-only successes: ten under the original instruction and twelve under revised wording. Takeover occurs in the other 15 gains and in nine of the 11 losses. Across the 200 episodes, takeover accounts for 21,182 of 227,198 executed control steps (9.3\%), including base and torso motion. The frozen VLA generates the remaining 90.7\%. Since takeover generally follows a stalled or incorrect action, these counts alone do not measure its effect.

\begingroup
\fontsize{9.2}{11.2}\selectfont
\setlength{\tabcolsep}{4pt}
\renewcommand{\arraystretch}{1.14}
\begin{longtable}{@{}P{0.235\linewidth}C{0.17\linewidth}C{0.17\linewidth}C{0.175\linewidth}C{0.175\linewidth}@{}}
\caption{Highest level of intervention used with Astra in 200 paired episodes. Percentages are computed within each outcome column.}\label{tab:intervention-by-outcome}\\
\toprule
\textbf{Highest intervention} & \textbf{Harness only\newline $(n=37)$} & \textbf{Policy only\newline $(n=11)$} & \textbf{Both succeed\newline $(n=104)$} & \textbf{Both fail\newline $(n=48)$} \\
\midrule
\endfirsthead
\toprule
\textbf{Highest intervention} & \textbf{Harness only\newline $(n=37)$} & \textbf{Policy only\newline $(n=11)$} & \textbf{Both succeed\newline $(n=104)$} & \textbf{Both fail\newline $(n=48)$} \\
\midrule
\endhead
\bottomrule
\endfoot
\bottomrule
\endlastfoot
Observation only & 27.0\% & 0.0\% & 88.5\% & 4.2\% \\
\addlinespace[2pt]
Instruction rewrite only & 32.4\% & 18.2\% & 6.7\% & 20.8\% \\
\addlinespace[2pt]
Any takeover & 40.5\% & 81.8\% & 4.8\% & 75.0\% \\
\end{longtable}
\endgroup

We therefore looked for three events within each harness-only success: the policy stalled on a specific substage, the VLM intervened to address it, and the robot subsequently completed that same substage. Twenty-three episodes meet this criterion: 12 with language alone and 11 with takeover. In one, Xiaomi touched the faucet and withdrew with no water visible. After Astra narrowed the instruction to “Turn on the sink faucet,” the cameras showed running water. Ten gains keep the original instruction and involve no takeover. Fixed reissue, which uses no VLM, also solves nine of them (\S\ref{sec:ablations}), so we do not credit them to the VLM's decisions. Four remain unresolved: three use takeover before any policy stall, and in one the policy succeeded on a later retry after both the rewrite and the attempted regrasp failed.

\noindent\begin{minipage}{\linewidth}
\begin{figure}[H]
  \centering
  \includegraphics[width=0.62\linewidth]{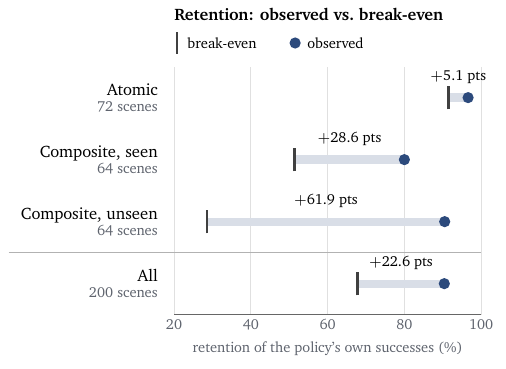}
  \caption{Retention of Xiaomi-Robotics-1 successes under GPT-6 Astra supervision. Dots show the fraction of standalone successes retained. Vertical marks show the minimum retention needed to break even at each group's observed rescue rate. Labels above the lines give the difference in percentage points. Scene counts give the total number of pairs in each group.}
  \label{fig:retention-margin}
\end{figure}

At the observed rescue rates, Eq.~(\ref{eq:retention}) gives a break-even retention of 91.5\% on atomic tasks, against 96.6\% observed (Figure~\ref{fig:retention-margin}). Three additional losses, a drop of just over five percentage points in retention, would erase the atomic gain. Seen and unseen compositions have wider margins: retention reaches 80.0\% against a 51.4\% threshold on seen tasks and 90.5\% against 28.6\% on unseen tasks. Seen compositions nevertheless account for seven of the 11 losses.
\end{minipage}

Switching from Sol to Astra raises rescue from 27.1\% to 43.5\%, while retention rises only from 89.6\% to 90.4\%. Fourteen of Astra’s 15 net additional successes thus come from scenes Xiaomi alone misses. Overall, the gain comes mainly from interventions: 27 of the 37 gains involve a rewritten instruction or a takeover, and 23 show evidence that the intervention helps.

\subsection{Ablations}

\label{sec:ablations}

Does the gain come from the VLM's decisions, or would running the same policy through the harness a second time yield a similar result? On the same 200 scenes, we ran the frozen Xiaomi policy through the same execution code and reissued the original task instruction every 160 control steps. This control makes no VLM calls, never rewrites the instruction, and leaves every action to the policy. Of the three design choices in \S\ref{sec:setup}, it keeps the default of the first, giving the policy the full instruction, and removes the other two: nothing judges progress and nothing intervenes.

Fixed reissue solves 113/200 scenes (56.5\%), close to the 115/200 solved by Xiaomi alone (Table~\ref{tab:supervisor-control}). In a direct comparison, Astra solves 30 scenes that fixed reissue misses, while fixed reissue solves two that Astra misses. The corresponding counts are 6 to 0 on atomic tasks, 12 to 0 on seen compositions, and 12 to 2 on unseen compositions.

\begin{table}[htbp]
\centering
\caption{Successes on the same 200 Xiaomi scenes, grouped by whether the standalone policy succeeds or fails.}
\label{tab:supervisor-control}
\small
\begin{tabular}{@{}lrrr@{}}
\toprule
Method & Total (/200) & \shortstack[r]{Policy succeeds\\(/115)} & \shortstack[r]{Policy fails\\(/85)} \\
\midrule
Xiaomi alone              & 115 & 115 &  0 \\
Fixed reissue (no VLM)    & 113 &  96 & 17 \\
Xiaomi + harness (Sol)    & 126 & 103 & 23 \\
Xiaomi + harness (Astra)  & 141 & 104 & 37 \\
\bottomrule
\end{tabular}
\end{table}

Astra solves 22 of the 68 scenes that both Xiaomi alone and fixed reissue miss (32.4\%) and retains all 96 that both solve. Of Astra's 37 gains over Xiaomi alone, fixed reissue reproduces nine of ten observation-only gains, three of twelve gains with rewriting alone, and three of fifteen gains involving takeover. Fixed reissue thus fails to reproduce 22 of the 37 gains, and 21 of those 22 involve an intervention: nine rewrite the instruction and 12 involve takeover. These gains therefore depend on judging progress and intervening, the two design choices that the control removes. All 11 scenes that Xiaomi solves alone but loses with Astra also fail under fixed reissue, so the paired loss alone cannot establish that an intervention caused the failure.

Fixed reissue achieves 56.5\% success. The two VLM configurations achieve 63.0\% with Sol and 70.5\% with Astra. Of the 85 scenes that Xiaomi fails alone, fixed reissue solves 17, Sol 23, and Astra 37.

We did not test a stall-triggered retry without a VLM or repeat the standalone policy run on the same machine as the fixed-reissue control, nor did we individually remove rewriting, takeover, or calibration. Within these limits, the gain needs the VLM: the same harness without one solves two fewer scenes than the policy alone and reproduces only 15 of the 37 gains.

\subsection{Failure analysis}

\label{sec:failure-analysis}

Where does supervision break down when Xiaomi loses a task it solves alone or an intervention leaves a substage unfinished?

Astra intervenes in all 11 lost successes (Table~\ref{tab:paired-loss-events}). Seven occur on seen compositions, including three in which a corrective grasp misses and the takeover tool then keeps the gripper closed, so no second attempt is possible. This is a limitation of the tool's implementation rather than of the supervision rules. In a possible false stall alert, Astra instructed Xiaomi to retrieve a spatula the robot was already using to stir corn.

We classified the 85 standalone failures using Section~\ref{sec:failures} before examining paired outcomes. With Astra, Xiaomi solves 11/29 scenes with decision failures (37.9\%) and 24/50 with execution failures (48.0\%). Rescue rates are similar for order or progress errors (55.6\%) and inconsistent execution (55.0\%), and lower for wrong targets (30.0\%) and execution failures on tasks with no observed standalone success (20.0\%). Six failures have indeterminate causes (Table~\ref{tab:rescue-by-failure-type}).

\begingroup
\fontsize{9.2}{11.2}\selectfont
\setlength{\tabcolsep}{4pt}
\renewcommand{\arraystretch}{1.14}
\begin{longtable}{@{}P{0.56\linewidth}C{0.22\linewidth}C{0.16\linewidth}@{}}
\caption{Paired rescue rate by the primary error in the standalone Xiaomi run. Each rate counts scenes completed with Astra among the standalone failures of that type.}\label{tab:rescue-by-failure-type}\\
\toprule
\textbf{Standalone failure type} & \textbf{Rescued / failures} & \textbf{Rescue rate} \\
\midrule
\endfirsthead
\toprule
\textbf{Standalone failure type} & \textbf{Rescued / failures} & \textbf{Rescue rate} \\
\midrule
\endhead
\bottomrule
\endfoot
\bottomrule
\endlastfoot
Wrong target & 6/20 & 30.0\% \\
\addlinespace[2pt]
Order or progress error & 5/9 & 55.6\% \\
\addlinespace[2pt]
Inconsistent execution & 22/40 & 55.0\% \\
\addlinespace[2pt]
Execution failure on a task with no observed standalone success & 2/10 & 20.0\% \\
\addlinespace[2pt]
Indeterminate & 2/6 & 33.3\% \\
\addlinespace[2pt]
\textbf{Overall} & \textbf{37/85} & \textbf{43.5\%} \\
\end{longtable}
\endgroup

Of the 48 scenes neither system solves, 35 trigger a stall alert and 46 receive an intervention. An intervention does not require an alert: a dropped object, an unsafe motion, or a confirmed target conflict can trigger one earlier (\S\ref{sec:language}). The final obstacle involves execution in 25 scenes and decisions in 13, with ten unresolved. This 35/48 is alert coverage, a measure of how often a warning appears. Estimating recall requires independent labels for all actual stalls.

Recovery after an intervention remains a weakness. In the Sol trajectory audit (Table~\ref{tab:episode-sets}), the harness recovers 11 of 35 failed skill attempts after intervention, with 22 unfinished and two indeterminate. Of these 35, 25 involve a takeover: nine recover, 14 remain unfinished, and two are indeterminate. The audit was carried out on the Sol runs, which were completed before the Astra evaluation.

In the main evaluation with Astra, 35 first interventions immediately follow a recorded error. The affected substage later succeeds in 20 cases, fails in 13, and remains indeterminate in two. Safety rules block six retries or handoffs, the policy fails to find or select the target after five handoffs, and two dropped objects remain unrecovered. Supervision therefore breaks down after it intervenes, not before: every lost success and 46 of the 48 scenes that neither run solves involve an intervention, and what fails is the corrective grasp, the retry, or the handoff that follows (\S\ref{sec:discussion}).

\section{Discussion}
\label{sec:discussion}

The paired evaluations show that the same supervision rules can improve several frozen policies while preserving most of their standalone successes. The practical value of this approach depends on how much of the benefit carries over when the policy changes, how reliably the robot recovers after an intervention, and when the gains justify the added latency of VLM supervision.

\noindent\textbf{What transfers across policies.}
On the same 50 scenes, supervision adds eight to ten percentage points to each of five frozen policies and preserves 58 of the 59 standalone successes across the four new policies. Transfer requires a native adapter and a policy skill profile. We prepared no task memories, exploratory rollouts, demonstrations, or scripts. What carries over less well is rescue: the harness recovers 18 of 141 standalone failures on the four new policies, against six of 20 for Xiaomi on the same scenes. The gain from per-task preparation, in our test of Harness VLA, did not carry over to the stronger policy (\S\ref{sec:harness-vla-transfer}).

\noindent\textbf{Physical recovery remains difficult.}
The VLM can change what a frozen policy attempts and when it attempts it, but the policy executes each grasp or placement with the motor skills it already has. When the policy lacks the skill, the remaining option is takeover, and takeover generally follows a harder failure: across all 200 scenes, 20/65 episodes involving takeover succeed, compared with 19/31 involving rewriting alone (Table~\ref{tab:intervention-by-outcome}). A takeover also ends with a handoff, after which the policy must continue from a pose chosen by the VLM; the failed recoveries in \S\ref{sec:failure-analysis} include five in which the policy cannot find or select the target after the handoff. Whether future VLAs or foundation models learn to recover on their own remains open. For the frozen policies tested here, the VLA alone still leaves recoverable failures unresolved.

\noindent\textbf{When supervision is worth its cost.}
Xiaomi gains only four percentage points on atomic tasks. For a policy that already succeeds 95\% of the time, even rescuing 43.5\% of its failures requires retaining about 98\% of its successes to break even. Supervision averages 71 VLM requests per episode, and waiting for responses consumes 87\% of episode wall time. Given that most runtime goes to VLM requests, which decisions need the stronger model? Sol and Astra retain nearly the same number of standalone successes but rescue 23 and 37 failures (\S\ref{sec:gain}). This suggests a division of labor in which a smaller model monitors routine progress and calls a stronger model when a substage needs recovery. Supervision may also become more valuable on longer tasks: Xiaomi's gains rise from four points on atomic tasks to 16 on seen compositions and 20 on unseen compositions, yet an episode averages only 57 seconds of simulated motion, and the longest lasts 3.6 minutes. Longer tasks offer more opportunities for a recoverable error: if each substage succeeds 90\% of the time, a task of 50 substages finishes only 0.5\% of the time unless failed substages can be recovered.

\noindent\textbf{Limitations.}
We developed the harness with Xiaomi, so the higher rescue rate on Xiaomi could reflect either Xiaomi's capabilities or design choices informed by that policy. Our transfer evidence covers 50 simulated scenes per new policy and one competing harness. We did not compare with EmbodiedRSI~\citep{embodiedrsi2026}, which learns skills and memory during exploration. We have yet to test whether a smaller model would escalate early enough to preserve Astra's gains. The trend across task categories is an extrapolation because the three categories differ in familiarity as well as complexity; testing it calls for longer, previously unseen compositions with recovery scored separately at each failed substage.

\section{Conclusion}
\label{sec:conclusion}

Our harness gives a frozen VLA the full task and monitors progress at each substage, responding to stalls by revising the instruction and, when needed, briefly taking over. On 200 RoboCasa365 scenes, supervision raises Xiaomi-Robotics-1's success rate from 57.5\% to 70.5\%, compared with 56.5\% when the harness reissues the original instruction without a VLM. \ours{} meets the first two properties of an ideal harness by design and comes close to the third. It needs no preparation for a new task: the VLM forms its plan from the instruction and the scene in each episode. It needs no redesign for a new policy: the same supervision rules add eight to ten percentage points to each of five frozen policies on the shared 50 scenes. It breaks little that a policy already solves, preserving 104 of Xiaomi-Robotics-1's 115 standalone successes and 58 of 59 on the four other policies. The gain therefore survives a change of policy, whereas a harness prepared for each task does not improve the stronger policy in our test (\S\ref{sec:harness-vla-transfer}). Rescue rates vary with the capabilities of both the policy and the VLM, and physical recovery remains the main bottleneck. Our next steps are to improve recovery after failed grasps and placements, test a smaller VLM for routine monitoring, and evaluate the harness on longer tasks and on physical robots.

\bibliography{references}
\bibliographystyle{plainnat}

\clearpage
\appendix
\counterwithin{table}{section}
\section{Detailed results on RoboCasa365}
\label{app:paired-results}

\begingroup
\fontsize{9.2}{11.2}\selectfont
\setlength{\tabcolsep}{4pt}
\renewcommand{\arraystretch}{1.14}
\begin{longtable}{@{}P{0.235\linewidth}C{0.15\linewidth}C{0.175\linewidth}C{0.19\linewidth}C{0.175\linewidth}@{}}
\caption{Paired outcomes for Xiaomi alone and Xiaomi with Astra. Entries are percentages within each row.}\label{tab:paired-outcomes}\\
\toprule
\textbf{Task category} & \textbf{Both succeed} & \textbf{Only Xiaomi alone succeeds} & \textbf{Only Xiaomi with Astra succeeds} & \textbf{Neither succeeds} \\
\midrule
\endfirsthead
\toprule
\textbf{Task category} & \textbf{Both succeed} & \textbf{Only Xiaomi alone succeeds} & \textbf{Only Xiaomi with Astra succeeds} & \textbf{Neither succeeds} \\
\midrule
\endhead
\bottomrule
\endfoot
\bottomrule
\endlastfoot
Atomic-seen & 79.2\% & 2.8\% & 6.9\% & 11.1\% \\
\addlinespace[4pt]
Composite-seen & 43.8\% & 10.9\% & 26.6\% & 18.8\% \\
\addlinespace[4pt]
Composite-unseen & 29.7\% & 3.1\% & 23.4\% & 43.8\% \\
\addlinespace[4pt]
Overall & 52.0\% & 5.5\% & 18.5\% & 24.0\% \\
\addlinespace[4pt]
\end{longtable}
\endgroup

\begingroup
\fontsize{9.2}{11.2}\selectfont
\setlength{\tabcolsep}{4pt}
\renewcommand{\arraystretch}{1.14}
\begin{longtable}{@{}P{0.17\linewidth}P{0.29\linewidth}C{0.11\linewidth}P{0.36\linewidth}@{}}
\caption{Failure categories in the Xiaomi-Robotics-1 audit.}\label{tab:failure-categories}\\
\toprule
\textbf{Failure category} & \textbf{Definition} & \textbf{Share of failures} & \textbf{Example} \\
\midrule
\endfirsthead
\toprule
\textbf{Failure category} & \textbf{Definition} & \textbf{Share of failures} & \textbf{Example} \\
\midrule
\endhead
\bottomrule
\endfoot
\bottomrule
\endlastfoot
Execution with no observed task success & The robot attempts the required action on the correct target but fails physically. The policy records no successes in at least five trials of the task. & 15\% & The gripper repeatedly tried to lift toast from the toaster. Both slices remained in their slots. The policy completes the task in 0 of 5 trials. \\
\addlinespace[4pt]
Inconsistent execution & The robot attempts the required action on the correct target but fails physically on a task it has completed in other trials. & 38\% & The robot placed a pitcher inside the cabinet, but it tipped and fell to the floor after release. The policy completes the task in 11 of 15 trials. \\
\addlinespace[4pt]
Wrong target & The robot acts on an object, appliance, or control other than the one specified by the instruction. & 24\% & Asked to move a chicken drumstick to the freezer, the robot picked up a banana. The drumstick remained on the refrigerator shelf. \\
\addlinespace[4pt]
Order or progress error & The robot skips a required step, advances before a step is complete, or repeats a step it has already completed. & 19\% & Asked to heat both bread and a kebab, the robot put the bread in the toaster oven and shut the door while the kebab was still on the plate. \\
\addlinespace[4pt]
\end{longtable}
\endgroup

\begingroup
\fontsize{9.2}{11.2}\selectfont
\setlength{\tabcolsep}{4pt}
\renewcommand{\arraystretch}{1.14}
\begin{longtable}{@{}P{0.65\linewidth}C{0.12\linewidth}C{0.17\linewidth}@{}}
\caption{Events observed in the 11 episodes that Xiaomi solves alone and loses with Astra. Each episode contributes to one row.}\label{tab:paired-loss-events}\\
\toprule
\textbf{Observed event} & \textbf{Episodes} & \textbf{Share of losses} \\
\midrule
\endfirsthead
\toprule
\textbf{Observed event} & \textbf{Episodes} & \textbf{Share of losses} \\
\midrule
\endhead
\bottomrule
\endfoot
\bottomrule
\endlastfoot
Observation only & 0 & 0.0\% \\
\addlinespace[2pt]
Instruction rewritten during observable progress & 1 & 9.1\% \\
\addlinespace[2pt]
Confirmed state damage from takeover & 0 & 0.0\% \\
\addlinespace[2pt]
Initial takeover consumed about half the step budget & 2 & 18.2\% \\
\addlinespace[2pt]
Plan omitted a required repair & 1 & 9.1\% \\
\addlinespace[2pt]
Missed corrective grasp after which the takeover tool kept the gripper closed & 3 & 27.3\% \\
\addlinespace[2pt]
Earlier policy failure or unresolved attribution & 4 & 36.4\% \\
\end{longtable}
\endgroup

\section{Evaluation protocol details}
\label{app:evaluation-protocol}

\subsection{Episode selection and planning timeouts}

The main evaluation used the Xiaomi-Robotics-1-RoboCasa365 checkpoint (0d1aa76d) and harness v1.2.3.5 (c8911c64). It ran on November 3–7, 2026, with a 330-second API timeout and a 180-second default planning timeout. RLDX-1 used the \texttt{RLDX-1-FT-RC365} checkpoint (c53d0bc0), and WorldDreamer used the \texttt{world\_dreamer-robocasa365-multi\_task} checkpoint (e570a06b). The primary Xiaomi analysis used episode indices 1--4. After outcomes for the 50 index-0 scenes became available, we decided to exclude them from the primary summary. All 50 have verified environment scores. The harness solves 25 and Xiaomi alone solves 24.

Ten initially unsuccessful Xiaomi--Astra runs ended early with planning timeouts. We reran those scenes with a 600-second planning timeout in place of the 180-second default. These were all the runs that ended on a planning timeout; such a run is cut short and can only be scored as a failure, so the reruns were not selected by outcome. Two changed from failure to success. The harness thus solves 141/200 scenes with the reruns and 139/200 using the original scores, compared with 115/200 for Xiaomi alone. Including index 0 gives 166/250 versus 139/250 with the reruns, or 164/250 versus 139/250 using the original scores. The separate 50-scene evaluations of the four other policies used a 600-second planning timeout and a 330-second API timeout.

\subsection{Stall windows used for each policy}
\label{app:stall-windows}

RLDX-1 and WorldDreamer use stall windows derived from their own profiling rollouts (\S\ref{sec:calibration}). We did not derive policy-specific windows for GR00T N1.5 and \(\pi_{0.5}\), which use the default windows: per-skill values from 62 to 476 control steps and a 299-step fallback for substages with no matching skill. Xiaomi-Robotics-1, the development policy, uses a single 120-step window.

\subsection{Harness VLA comparison protocol}

Table~\ref{tab:harness-vla-transfer}, Panel~A, uses the success rates tabulated in arXiv v5 of \citet{harnessvla2026}. Appendix~C.3, Table~12 of that paper specifies five evaluation seeds per task (250 rollouts), and Appendix~H reports low reasoning effort for GPT-6 Astra. The authors' \href{https://github.com/RLinf/RPent/blob/f2e6b1dc5ee61e16bfaef7789eeb40bb31673ac3/docs/_static/leaderboard/results.json#L877-L907}{released evaluation record} lists 340 rollouts (180 atomic, 80 seen composite, and 80 unseen composite) and documents a correction of the Astra reasoning effort to high. Both sources report 59.2\% overall for Astra.

\subsection{Failure audit}
\label{app:failure-audit}

The audited failures come from an earlier RoboCasa365 evaluation of Xiaomi-Robotics-1 running without a harness, which produced 2,413 valid rollouts, 893 of them failures. We sampled 193 failures across five task groups (the atomic-seen, composite-seen, and composite-unseen tasks evaluated in \S\ref{sec:protocol}, all 65 atomic tasks, and tasks that appear only in the pretraining set), selecting up to 50 episodes per group and at most three per task. For each episode, we reviewed the instruction alongside synchronized keyframes from three cameras and simulator records of object motion, gripper contact, and appliance state. We labeled the earliest observable error that diverted the robot from the task. To check consistency, we reannotated 45 episodes. Of the 44 episodes for which we assigned a definite label on the second pass, 38 received the same primary label both times (86\%).

\subsection{Episode sets and their overlap}
\label{app:episode-sets}

The shared 50-scene subset is contained in the main evaluation. Of the 80 episodes in the Sol trajectory audit, 64 use main-evaluation scenes (episode indices 1--4 of the 16 unseen-composition tasks) and 16 use episode index 0. The audit reads trajectories supervised by Sol, whereas the main results use Astra. Of the 72 development episodes, 28 trace to failures included in the failure audit, and 10 match main-evaluation scenes on the task, the initial simulator state, and the starting camera images (\S\ref{sec:protocol}).

\subsection{What the harness knows about a task}
\label{app:task-information}

We count as preparation anything made for one task that tells a harness how to solve it: demonstrations, exploration episodes on that task, stored memories, or scripts. \ours{} uses none of these. Its rules and prompts contain nothing written for a particular task, and it keeps no memory between episodes. The only prior information is the skill profile, which records how often and how quickly the policy alone completes each skill and is measured without any VLM call (\S\ref{sec:calibration}). The profile tells the harness how much to trust the policy, not how to solve the task, and a skill without an entry uses policy-wide defaults. This information overlaps with the evaluated tasks in three ways. The 65 profiled atomic tasks include the 18 atomic tasks we evaluate. Five of the 32 evaluated composite tasks also have a whole-task entry in the profile, covering 20 of the 128 composite episodes. Finally, we developed the rules on 72 episodes that cover 47 of the 50 tasks (\S\ref{sec:protocol}).

\subsection{Example episodes}
\label{app:development-cases}

\noindent\textbf{An episode without intervention (\S\ref{sec:capability}).} In one episode, the robot was asked to add a lemon wedge and an ice cube to a glass of lemonade. At steps 160, 320, and 480, the VLM recorded that the gripper had reached the refrigerator handle, the door had opened, and the gripper held the lemon wedge. It continued to check every 160 steps through step 960. The policy completed the task at step 1,000 under the same instruction it had received at the start.

\noindent\textbf{Handing over the whole task (\S\ref{sec:capability}).} We adopted this rule after observing an early failure on a tea-serving task. An earlier version split the task into moving a kettle, placing a mug on the same tray, and closing the cabinet doors. It took over to move the kettle and later instructed the policy to close the doors while the mug was still off the tray. On the same scene, the policy completed the full sequence in 1,415 steps when given the original instruction from the outset.

\noindent\textbf{Judging progress at the current substage (\S\ref{sec:progress}).} A matched development run illustrates why the scope of the progress signal matters. The robot had to place a loaf of bread in a basket and then carry the basket to the dining counter. The earlier version of the harness repeatedly credited the robot for holding the loaf as it drove through the kitchen. It was still reporting progress after the basket had disappeared from view. By the time it flagged a placement failure at step 832, the robot had released the loaf onto a bare counter. The two versions executed identical actions through step 320. At that checkpoint, the current version found no evidence that the loaf was getting closer to the basket and prompted the policy to “Place the bread in the basket on the counter.” At step 512, the wrist camera showed the loaf inside. The harness then advanced to moving the basket, and the environment registered success at step 1,127.

\subsection{Takeover constants}
\label{app:constants}

A takeover's initial allowance is 1.5 times the skill’s measured typical duration, or 500 control steps when no measurement is available. The harness ends the attempt after 150 executed steps without progress or upon a verified drop or unsafe motion. If the robot is already moving a securely held object toward placement, it can receive up to 50 additional steps at a time, capped at 200 per operation.

\subsection{Harness VLA transfer evaluation}
\label{app:harness-vla-transfer}

Our transfer evaluation used one scene per task and GPT-6 Astra at \texttt{xhigh} reasoning effort. The published evaluation used multiple seeds per task (Appendix~\ref{app:evaluation-protocol}). In the released memory set, four of the 50 tasks (OpenCabinet, OpenDrawer, HeatKebabSandwich, and WashFruitColander) lack the task-specific Markdown notes but keep their other task-specific files and the global memory. We evaluated all four using the remaining released memory. Harness VLA and our harness each solve three of these four scenes, and Xiaomi alone solves two. On the remaining 46 scenes, Harness VLA solves 26, Xiaomi alone 28, and our harness 31.

None of the four scenes that Xiaomi solves alone and Harness VLA loses (CoffeeSetupMug, GetToastedBread, KettleBoiling, and WaffleReheat) is among these four tasks. Both runs had the same control-step limit in each scene. Xiaomi alone first succeeds at steps 209, 1,885, 1,211, and 1,216, while the Harness VLA runs use the full limits of 600, 3,000, 1,500, and 4,050 steps without completing the task. They stop at different points: the mug is set down outside the valid position under the dispenser, the toast is never taken out of the toaster, the kettle is left 0.30\,m from the burner, and the microwave is never switched on.

\end{document}